\documentclass{article}

\usepackage{multirow}
\usepackage{pifont}
\usepackage{makecell}
\usepackage[dvipsnames]{xcolor}
\usepackage{adjustbox}

\newcommand{\cmark}{\ding{51}} 

\newcommand{\xmark}{\ding{55}} 

 \usepackage[main, final]{neurips_2026}

\usepackage[utf8]{inputenc} 
\usepackage[T1]{fontenc}    
\usepackage{hyperref}       
\usepackage{url}            
\usepackage{booktabs}       
\usepackage{amsfonts}       
\usepackage{nicefrac}       
\usepackage{microtype}      
\usepackage{xcolor}         
\usepackage{enumitem}
\usepackage{graphicx}
\usepackage{subcaption}
\usepackage{amsmath,amssymb,amsthm,amsfonts,bm,bbm, nicefrac}
\usepackage{algorithm}
\usepackage{algpseudocode}
\usepackage{xcolor}         
\definecolor{mydarkblue}{rgb}{0,0.08,0.45}
\definecolor{DarkBlue}{rgb}{0,0,0.55}
\usepackage{graphicx}

\DeclareMathAlphabet{\mathsfit}{\encodingdefault}{\sfdefault}{m}{sl}
\SetMathAlphabet{\mathsfit}{bold}{\encodingdefault}{\sfdefault}{bx}{n}

\def\gA{{\mathcal{A}}}

\def\gD{{\mathcal{D}}}

\def\gO{{\mathcal{O}}}
\def\gP{{\mathcal{P}}}

\newenvironment{enumerate*}%
{\begin{enumerate}[leftmargin=*,topsep=0pt]%
		\setlength{\itemsep}{0pt}%
		\setlength{\parskip}{0pt}}%
	{\end{enumerate}}

\theoremstyle{plain}

\theoremstyle{definition}

\theoremstyle{remark}

\usepackage{array} 

\usepackage{subcaption}

\title{Dynamics-Informed Adaptive Offline RL for Real-Time Tokamak Plasma Control}

\author{%
  Rohit Sonker\thanks{equal contribution} \\
  Robotics Institute\\
  Carnegie Mellon University\\
  \And
  Xiaoyan Hu\footnotemark[1]\\
  Robotics Institute \\
  Carnegie Mellon University\\
  \AND
  Hiro Josep Farre Kaga \\
  Princeton University \\
  \& Princeton Plasma Physics Lab
  \And
  Andrew Rothstein \\
  Princeton University \\
  \And
  Jiayu Chen \\
  The University of Hong Kong \\
  \And
  Egemen Kolemen \\
  Princeton University \\ \& Princeton Plasma Physics Lab \\
  \And
  Jeff Schneider \\
  Carnegie Mellon University
}

\begin{document}

\maketitle

\begin{abstract}

Tracking plasma profiles, such as electron temperature and rotation, remains a challenging task in tokamak nuclear fusion. While RL-based approaches offer a promising path for multi-input-multi-output control systems, existing frameworks overlook two key obstacles:  partial observability of plasma dynamics and the diagnostic mismatch between offline training and online execution. We propose a novel offline RL framework to address these challenges. Specifically, we train the control policy to condition on the latent context extracted from a trained plasma dynamics model. This dynamics-informed latent context encodes information about operating regimes, enabling the policy to adapt as the plasma evolves. To mitigate the gap between offline and real-time observation, we learn a converter that maps offline reconstructed plasma diagnostics  to their real-time observable counterparts, thereby exposing policy training to the real-time plasma diagnostics available in live experiments. In offline evaluations, the proposed method substantially improves tracking performance and robustness across a diverse set of discharges. We report real-time deployment results on DIII-D Tokamak Fusion Facility, showing feasibility under challenging actuator conditions.




\end{abstract}









\section{Introduction}
Nuclear fusion is a promising path toward sustainable energy, and tokamaks are among the leading candidates for producing controlled thermonuclear fusion power. Conventional approaches to plasma control typically combine precomputed actuator trajectories, such as coil-current programs~\citep{Walker01112006}, with feedback controllers designed for specific target quantities. These control systems have demonstrated reliable performance within a wide range of discharges, but their design requires substantial engineering effort and expertise, especially for advanced plasma scenarios \citep{shi2017data}. One such example is the control of plasma profiles, including pressure, temperature, density, and rotation, where the control quantities are spatially varying rather than being a single scalar.


Recently, reinforcement learning (RL) has emerged as an alternative paradigm for building feedback controllers. In this paradigm, a control policy is first learned through interaction with a tokamak simulator and then deployed on the plasma control system. Existing RL-based approaches often rely on physics-based simulators, such as RAPTOR~\citep{felici2011} and Forward Grad-Shafranov Static~(FGE)~\citep{carpanese2021} simulator. These approaches have succeeded in generating novel plasma configurations~\citep{degrave2022magnetic, tracey2024towards} and avoiding disruptions in tokamak ramp-down~\citep{Wang2025}. However, physics-based simulators can be computationally expensive, especially during RL training, where agents repeatedly explore the actuator space and may encounter regimes that slow the convergence of iterative solvers. Additionally, these simulators are not capable of modelling kinetic profile evolution reliably \citep{abbate2024large}.  Finally, adapting such simulators to a specific tokamak often requires substantial parameter tuning and model calibration, thus making out of the box deployment on future tokamak devices difficult. In this work, we follow an alternative line of research that learns plasma dynamics directly from historical experimental discharges in a data-driven manner. These methods have demonstrated promising results in avoiding tearing instabilities~\citep{seo2024avoiding}, normalized plasma pressure control~\citep{char2023offline}, and rotation profile control~\citep{sonker2026offline}.

We highlight two challenges in existing RL approaches. First, existing RL-based policies typically condition the output actuators on a single-timestep observation of the plasma diagnostics. In practice, however, plasma evolution depends not only on the instantaneous state, but also on the operational phase, control scenario, and device conditions~\citep{sonker2025multitimescale}. As a result, such policies can fail to distinguish plasma regimes that appear similar at a given time but evolve differently thereafter. On the other hand, existing dynamics learning methods ~\citep{char2024full} use offline reconstructed diagnostics from EFIT~\citep{Lao_1985} and ZIPFIT \citep{logan2018omfit}, while the plasma control system instead receives low latency diagnostics deployment, such as real-time EFIT (RT-EFIT) and RTCAKENN~\citep{Shousha_2024}. These real-time diagnostics can be noisy and low-fidelity, which creates a substantial sim-to-real gap between training and deployment.

\begin{figure}[!t]
    \centering
    \includegraphics[width=1.\linewidth]{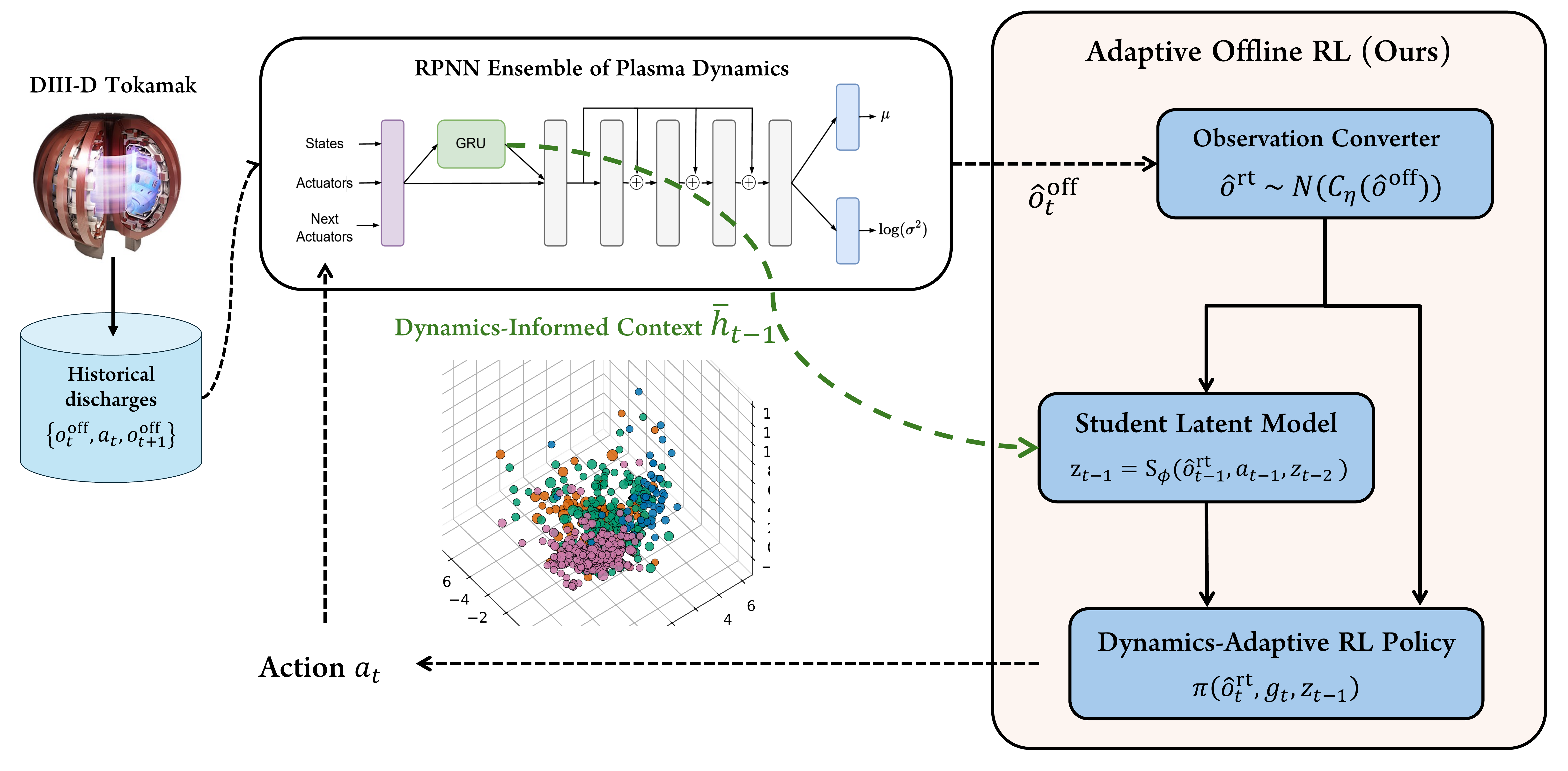}
    \caption{Adaptive offline RL framework: The student latent model is trained to predict dynamics-informed context that varies between plasma regimes. The observation converter exposes policy training to real-time tokamak diagnostics.}
    \label{fig:framework}
\end{figure}

In this work, we introduce a novel adaptive RL framework for fusion tokamak control~(See Figure~\ref{fig:framework}). To address partial observability in plasma dynamics, we condition the control policy on a latent context extracted from a recurrent probabilistic neural network~(RPNN)~\citep{char2024full}, which is trained on historical discharges in the DIII-D Tokamak to predict plasma evolution. 
We show that this dynamics-informed context varies with plasma operating phase and confinement mode (see Figure~\ref{fig:latent_phase_confinement}), and that conditioning on it improves policy learning.
Second, to mitigate the sim-to-real gap, we further introduce an observation converter that maps offline reconstructed diagnostics to their real-time-observable counterparts, thereby allowing policy training to better align with the tokamak environment during deployment. Results on RPNN and transformer-based simulated tokamak environments show that our proposed framework improves policy learning and yields stronger control performance against the baseline methods, including the single-step feedforward controller and the recurrent architecture-based policy. Finally, we evaluate the learned controllers on the DIII-D Tokamak. In summary, our contributions are as follows.

\begin{itemize}[leftmargin=*]

\item We propose a novel adaptive RL framework for plasma-profile control, where the control policy is conditioned on the latent context distilled from an RPNN-based plasma dynamics model.

\item We show that the learned latent context encodes useful plasma information, including shot-phase and confinement-regime structure, and improves policy learning over the baseline methods, including the feedforward controller and the recurrent architecture-based policy.

\item To address the sim-to-real gap, we introduce an observation adaptation pipeline that maps signals reconstructed offline to real-time diagnostics. This pipeline simulates real-time observations input to the policy and improves its robustness in a real-world tokamak environment.


\item We evaluate the learned controllers on the rotation and electron temperature ($T_e$) profile-control tasks in the DIII-D Tokamak to demonstrate the effectiveness of our proposed framework.

\end{itemize}

\section{Related Works}

\paragraph{Learning-based control in tokamaks.}

Reinforcement learning has recently been applied to tokamak control, with early success in magnetic shape control using physics-based simulators~\citep{degrave2022magnetic, tracey2024towards}. Extending these methods to profile control is harder because pressure, temperature, density, and rotation profiles are difficult to model accurately with first-principles simulators alone. Early work developed rotation profile controllers using linear quadratic control \citep{wehner2015toroidal} and Model Predictive Control \citep{wehner2017combined}, but these studies were limited to simulations. This has motivated data-driven approaches that learn plasma dynamics from historical experiments. \cite{char2023offline} used learned DIII-D dynamics models for offline model-based RL of normalized pressure and differential rotation, \citep{sonker2026offline} extending it to full profile control. Other recent works have studied learning-based feedback control for instability avoidance, current and shape control, and rampdown design~\citep{seo2024avoiding, wang2025learning}. Our work follows this data-driven direction, but uses the recurrent dynamics model not only as a simulator, but also as a source of adaptive context for the policy. 

\paragraph{Offline reinforcement learning.}

Offline RL learns policies from fixed datasets without new environment interaction~\citep{levine2020offline}, making it suitable for safety-critical systems such as tokamaks. Model-free offline RL methods such as CQL, IQL, and TD3+BC reduce the need for online exploration but can be difficult to apply when the dataset is small relative to the control complexity~\citep{kumar2020conservative, kostrikov2021offline, fujimoto2021minimalist}. Model-based RL instead learns a dynamics model and trains policies using synthetic rollouts. Probabilistic ensembles are commonly used to account for model uncertainty~\citep{chua2018deep}, and offline model-based methods such as MOPO, BA-MCTS,  restrict or penalize uncertain rollouts~\citep{yu2020mopo, chen2026bayes}. While offline RL is a promising direction, a comparison model-free offline RL methods and model-based offline RL methods for data-driven tokamak profile control was performed in \citep{sonker2026offline}. Therefore in this work, we focus on model-based offline RL, where a dynamics model is learned from the data and experience generated from it is used to train a policy.

\paragraph{Adaptive RL and context-aware dynamics.}

Adaptive RL methods address settings where the dynamics or task vary across episodes. Meta-RL methods such as RL$^2$ and MAML learn policies that can adapt from recent experience~\citep{duan2016rl2, finn2017model}, while latent-context methods such as PEARL and VariBAD infer compact task representations for adaptation~\citep{rakelly2019efficient, zintgraf2021varibad}. Online system-identification approaches similarly estimate hidden dynamics parameters and provide them to a universal policy~\citep{yu2017preparing}. The closest prior work to our method is CaDM, which learns a context-aware dynamics model for generalization in model-based RL~\citep{lee2020context}. CaDM shows that dynamics-relevant context can improve adaptation, so the basic idea of context-conditioned control is not new. Our distinction is in the tokamak setting: we repurpose the hidden state of a recurrent plasma dynamics model trained on historical shots and use it directly as policy context for profile control.
Thus, our motivation is to learn a single policy that adapts across multiple historical plasma scenarios without requiring scenario-specific identification or controller tuning.



\paragraph{Privileged information and observation adaptation.}

Many real-world RL systems use richer information during training than is available at deployment. Asymmetric actor-critic and teacher-student methods allow critics or teachers to use privileged state information while training actors or students that operate from deployable observations~\citep{pinto2018asymmetric, kumar2021rma}. We use these ideas in tokamak control because offline reconstructions such as CAKE, and ZIPFIT differ from real-time estimates such as RTEFIT and RTCAKENN~\citep{ xing2021cake, logan2018omfit, Shousha_2024}. Hence, we create an observation converter which allows our policy to learn with real-time-like signals and while modelling the simulation dynamics with offline high fidelity signals.




\section{Preliminaries}

\textbf{Modeling Plasma Control as a POMDP.}\quad We model tokamak profile control as a finite-horizon partially observable control problem with two observation domains, denoted by \(\mathcal{M}=(\mathcal{S},\mathcal{O}^{\mathrm{off}},\mathcal{O}^{\mathrm{rt}},\mathcal{A},\mathcal{T},\mathcal{G},R,\gamma,d_0)\). Here, \(\mathcal{S}\) is the underlying plasma state, \(\mathcal{O}^{\mathrm{off}}\) is the offline reconstructed observation space, \(\mathcal{O}^{\mathrm{rt}}\) is the real-time observation space, \(\mathcal{A}\) is the actuator space, \(\mathcal{T}:\mathcal{S}\times\mathcal{A}\rightarrow\mathcal{P}(\mathcal{S})\) is the transition kernel, \(\mathcal{G}\) is the target space, \(R:\mathcal{S}\times\mathcal{A}\times\mathcal{G}\rightarrow\mathbb{R}\) is the reward function, \(\gamma\) is the discount factor, and \(d_0\in\mathcal{P}(\mathcal{S})\) is the initial-state distribution. Since \(\mathcal{S}\) is not directly observed, the learned dynamics model is trained on reconstructed observations \(o_t^{\mathrm{off}}\in\mathcal{O}^{\mathrm{off}}\), which serve as a proxy state for model rollouts and reward computation. At time \(t\), the policy receives a real-time observation \(o_t^{\mathrm{rt}}\in\mathcal{O}^{\mathrm{rt}}\), a target \(g_t\in\mathcal{G}\), and optionally a lagged latent context \(z_{t-1}\), then selects \(a_t\sim\pi(\cdot\mid o_t^{\mathrm{rt}},g_t,z_{t-1})\). The state evolves as \(s_{t+1}\sim\mathcal{T}(\cdot\mid s_t,a_t,z_t)\). The objective is to learn a policy that maximizes \(\max_{\pi}\mathbb{E}[\sum_{t=0}^{H-1}\gamma^t R(s_t,a_t,g_t)]\), with \(s_0\sim d_0\). Detailed description of state and observation space is provided in Appendix \ref{apnd:rpnn}.

\textbf{Offline Model-Based RL with Data-Driven Plasma Dynamics Model.}\quad Due to safety and economic constraints, learning such a controller through online interaction in a real-world tokamak is impractical. We frame this problem as an offline model based RL problem where we first learn plasma dynamics using an \textit{offline} dataset denoted by $\gD := \{ (o_t, a_t, o_{t+1})\}$, which consists of $\sim$ 16K  historical experiment trajectories (shots) from the DIII-D Tokamak. These shots are typically 4-5 seconds long and are sampled at 20ms frequency (50 Hz). Our dataset consists of plasma profiles which measures the spatial variation of quantities such as electron temperature, pressure, rotation and density across time, and we aim to develop controllers for rotation and electron temperature profiles. Controlling these profiles provides several advantages to improve stability, confinement and plasma efficiency \citep{richner2024use, bardoczi2025tearing}. We process this data by converting 33D profiles to 4 PCA components which capture upto 99\% variance. Other scalar quantities and actuators including neutral beams, electron cyclotron heating (ECH) and shape control parameters are also used with full details mentioned in the Appendix \ref{apnd:rpnn}. This data spans over a decade of DIII-D experiments. Over these years, the tokamak has undergone multiple sensor, actuator and wall upgrades thus the inherent source data distribution keeps evolving.

\section{Adaptive Offline RL for Plasma Profile-Tracking}


Existing RL approaches for tokamak control \citep{seo2024avoiding, sonker2026offline}, typically condition the control policy on a single-step observation of the plasma diagnostics. However, because plasma evolution is closely related to the operating phase and control scenario, a single-step observation may not fully determine the future plasma response \citep{char2024full}. Additionally, since model dynamics is trained on offline reconstructions, deploying to real-time conditions creates a sim-to-real gap. In this section, we propose to address the challenges of partial observability and the sim-to-real gap through a model based offline RL framework. Our approach features two components: a dynamics-informed latent context for adaptive control and an observation converter that exposes the policy to real-time measurements during training. 


\begin{figure}[!t]
    \centering
    \begin{subfigure}[t]{0.42\linewidth}
        \centering
        \includegraphics[width=\linewidth]{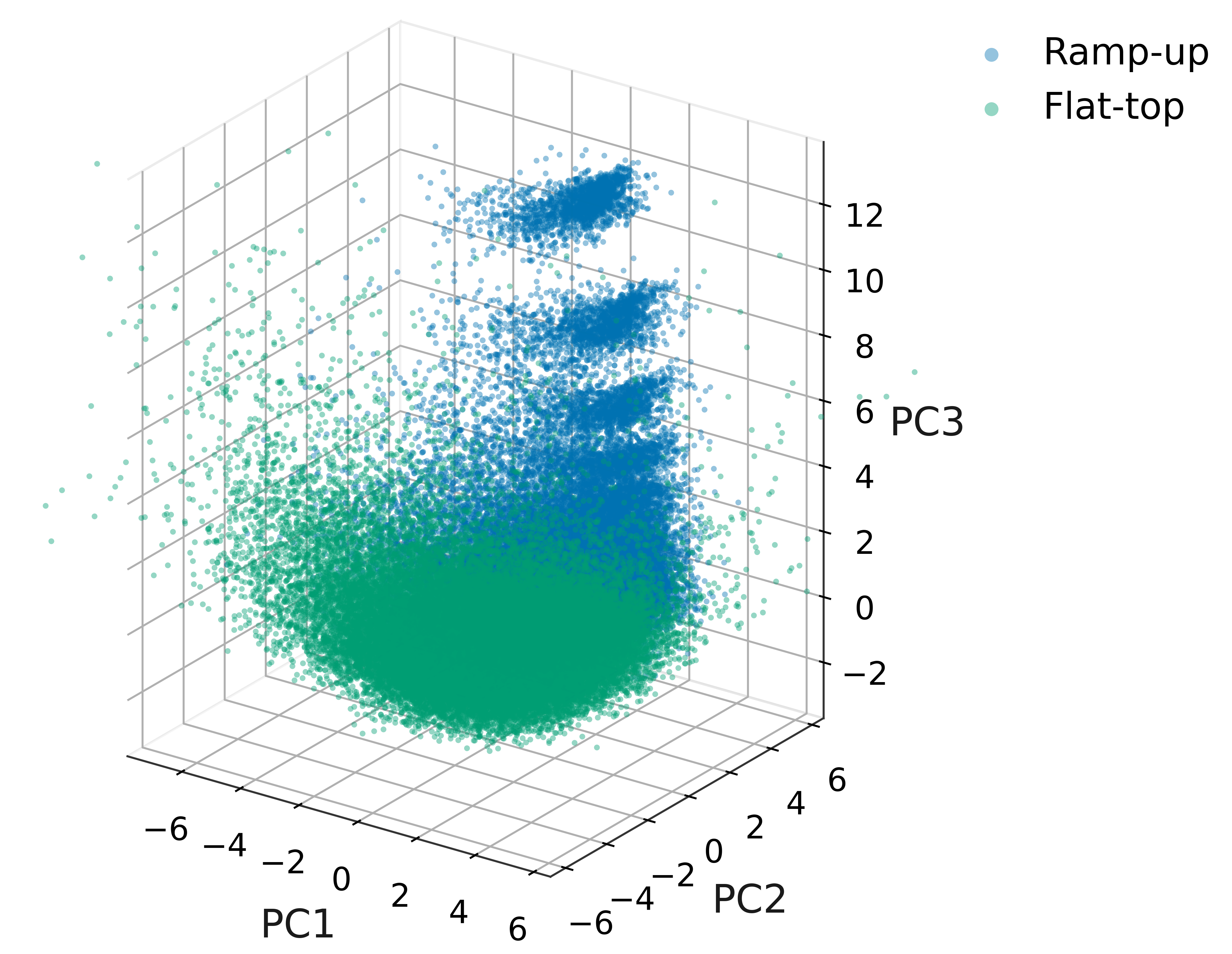}
        \caption{Latent space variation across plasma phases - Rampup and Flat-top}
        \label{fig:shot_phase}
    \end{subfigure}
    \qquad\qquad
    \begin{subfigure}[t]{0.42\linewidth}
        \centering
        \includegraphics[width=\linewidth]{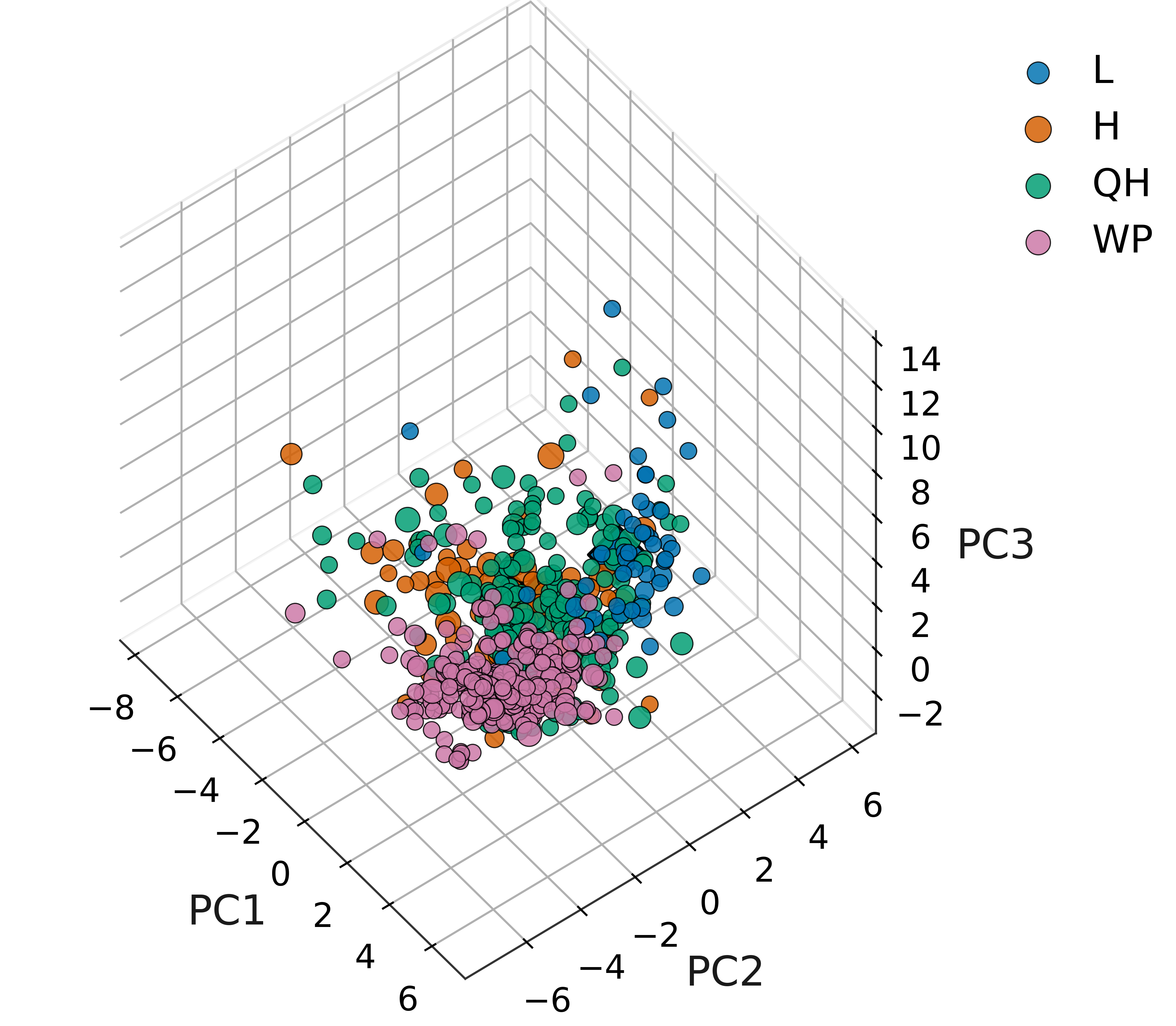}
        \caption{Latent space variation across plasma confinement modes - L mode, H mode, QH mode and WPQH mode}
        \label{fig:confinement_mode}
    \end{subfigure}

    \caption{Latent-space visualizations using Top 3 PCA components vs  phase and confinement mode labels.}
    \label{fig:latent_phase_confinement}
\end{figure}



\subsection{Dynamics-Informed Latent Context}

We first model the transition function as  $T_{\theta}: \gO^\text{off} \times \gA \to \gP (\gO^\text{off})$, parameterizing it by a bootstrapped ensemble of recurrent probabilistic neural networks~(RPNN)~\citep{char2024full}, which maps each observation-actuator pair to a distribution over the next observation. The RPNN contains a GRU recurrent unit and two output heads (a mean head and a log variance head). We train a bootstrapped ensemble of RPNNs; the predicted log variance captures aleatoric uncertainty, while disagreement across ensemble members provides an estimate of epistemic uncertainty. In particular, the observations, which we denote by $\gO^\text{off}$, are plasma diagnostics reconstructed offline known as ZIPFIT. They are smoother and therefore more suitable for long-horizon prediction than real-time diagnostics $\gO^\text{rt}$. This model is trained offline on a dataset of $\sim16$k shots. More training details are given in Appendix \ref{apnd:rpnn}.


Because the plasma state is partially observable, the RPNN's success is highly influenced by the presence of recurrent GRU layer. Two time points may have similar profile values and actuator settings but evolve differently because of underlying operating conditions or scenarios. Since next-state prediction provides a dense training signal at every timestep, the latent state is encouraged to encode information relevant to plasma evolution. Specifically, the latent context is updated from the recent trajectory, $z_t = f_{\phi}(s_t,a_t,z_{t-1})$ where $f_{\phi}$ is the recurrent latent model. Then, we train the control policy that takes actuator $a_t \sim \pi_{\psi}(a_t \mid o_t^{\mathrm{rt}}, g_t, z_{t-1})$. This is in contrast to using a recurrent policy, which learns its memory through the control objective, whereas $z_t$ is learned through next-state prediction. 

\subsection{Student Latent Model and Hidden State Distillation}
During policy learning, we use an ensemble of RPNN models, however, using all ensemble members to produce context during deployment is impractical. We therefore distill the ensemble hidden state into a single student recurrent model. For each timestep, we run the trajectory through every pretrained dynamics model and extract the recurrent hidden state $h_t^{(i)}$ from ensemble member $i$. These hidden states are averaged, $\bar{h}_t = \frac{1}{N}\sum_{i=1}^{N} h_t^{(i)}$, where $N$ is the number of ensemble members. We chose mean of latents as it is a low variance target, more details are in Appendix \ref{apd:observer}. A student GRU $S_\phi$ is then trained to predict $\bar{h}_t$ from the same input sequence by minimizing.

$$\mathcal{L}_{\mathrm{distill}}
=
\left\|
\hat{h}_t - \bar{h}_t
\right\|_2^2.
$$
The predicted hidden state $\hat{h}_t$ is used as the latent context $z_t$. This gives the policy access to the internal representation of the dynamics ensemble through a single efficient model.

When the observation converter is enabled $(C_{\eta})$, the student latent model is trained and updated using converted real-time observations $(\mathcal{O}^{\mathrm{rt}})$ as input. This matches the deployment setting, where the policy and latent model only receive real-time diagnostic estimates.

\subsection{Real-Time Observation Converter}
The dynamics model is trained in the offline reconstructed state space, but the deployed policy must act from real-time diagnostic estimates. To reduce this mismatch, we train an observation converter $C_{\eta}: \mathcal{O}^{\mathrm{off}} \rightarrow \mathcal{O}^{\mathrm{rt}}$, which maps ZIPFIT signals to RTCAKENN signals. We also attempted to directly train a dynamics model on the real-time signals, however, it was unsuccessful. This is primarily attributed to the higher noise and rapid variations in signal compared to offline reconstructions which are smooth.

During rollouts, we maintain the original state for dynamics prediction, reward computation, and critic training. The converted observation is used as input for the policy and as input for the latent state observer. The transition model remains in the offline state space, while the actor is trained on observations closer to the deployment distribution.

We parameterize the converter as a heteroscedastic neural network $o^\text{rt} \sim \mathcal{N}(\mu_\eta(o^\text{off}), \sigma_\eta(o^\text{off}))$. This heteroscedastic model predicts both the mean and uncertainty of the real-time observation. This model is trained offline by matching pairs of $(o^\text{off}_t, o^\text{rt}_t)$ data taken from $\sim5$k shots. Learning the uncertainty provides a better data fit. More details along with calibration metrics are provided in Appendix \ref{apd:converter}. This converter is used only at the observation level as a targeted domain shift projector and noise injector and does not change the state evolved by the learned dynamics model. Under this framework, the student model $(S_\phi)$ is also trained to predict context from the converted realtime outputs instead of true signals.

\begin{figure}[!t]
    \centering
    \begin{subfigure}[t]{0.48\linewidth}
        \centering
        \includegraphics[width=\linewidth]{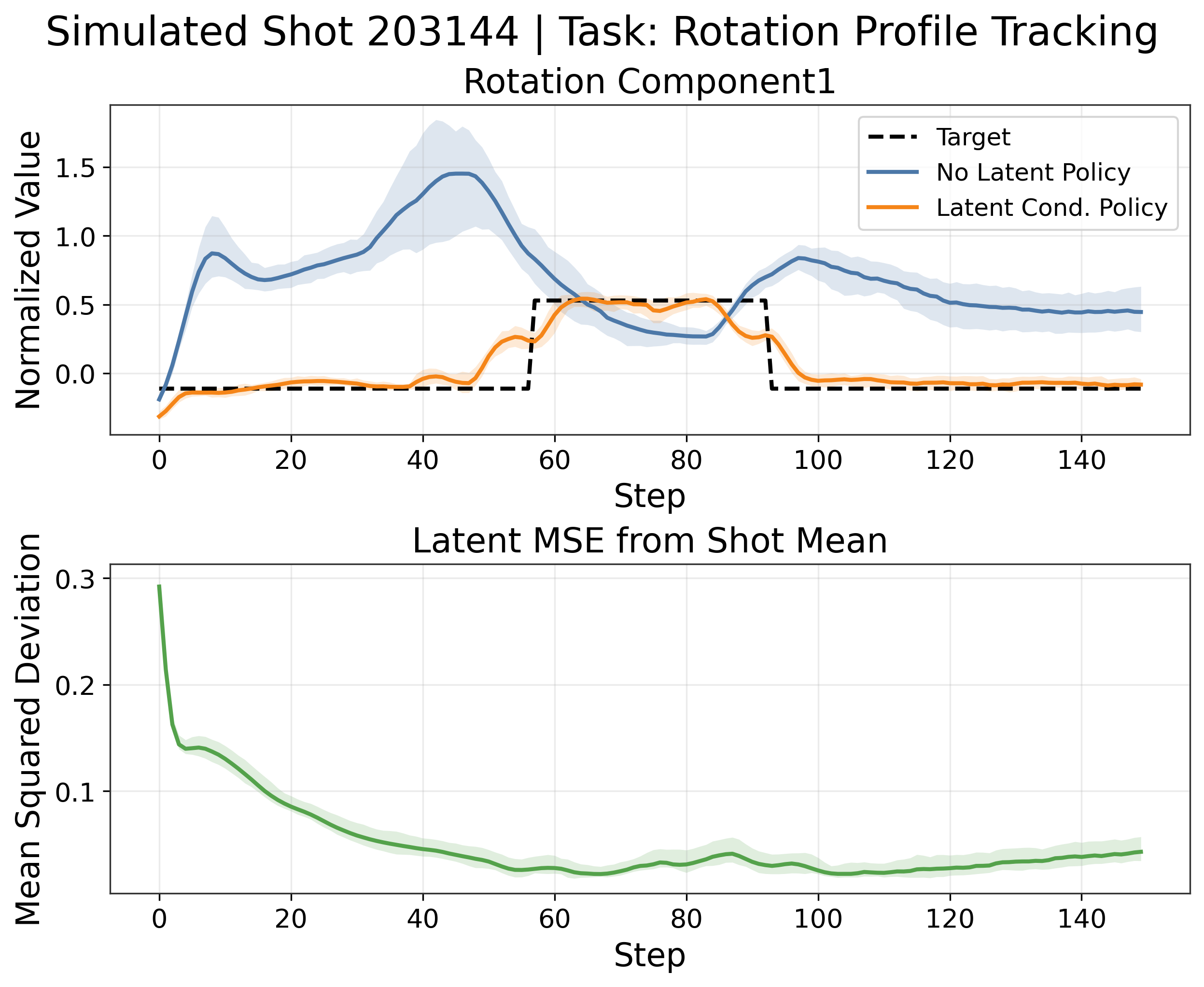}
        \caption{Rotation Profile control. We see latent variation at the start of the shot (ramp-up to flattop transition) and also towards end of shot, which affects tracking performance in No-latent policy }
        \label{fig:rot_variation}
    \end{subfigure}
    \hfill
    \begin{subfigure}[t]{0.48\linewidth}
        \centering
        \includegraphics[width=\linewidth]{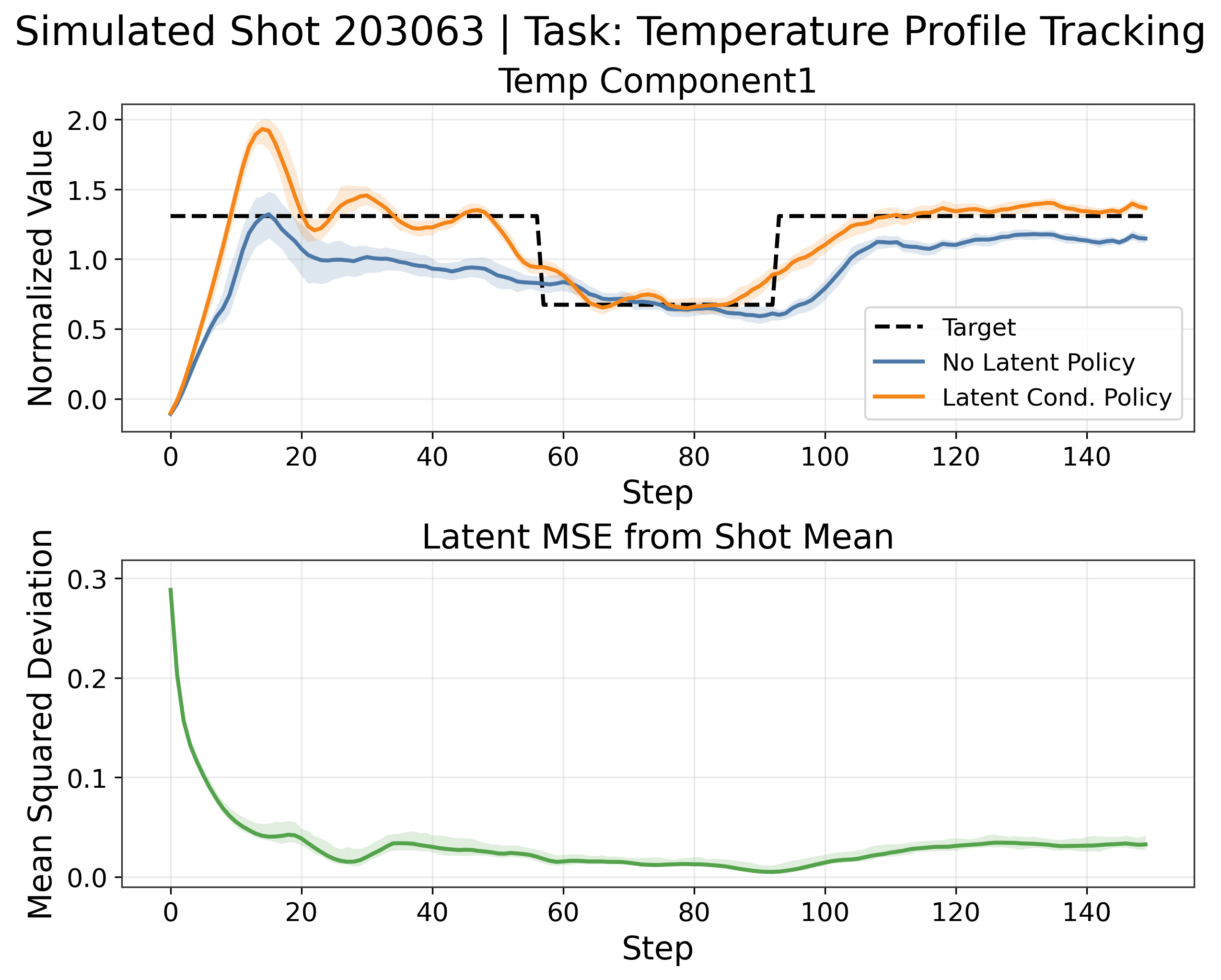}
        \caption{Electron Temperature Profile Control - Smaller impact due to startup (shorter ramp-up). Variations in latent space persist in shot, which lead to No-latent policy optimizing for average tracking performance.}
        \label{fig:temp_variation}
    \end{subfigure}

    \caption{Comparing Latent Conditioned vs No Latent Policies across Shot Roll-outs. Plots show tracking in first principal component of profile and variation in latent vector. Variation is measured as the MSE at each step from mean latent across the shot. Average over 30 rollouts, shaded region shows 95\% confidence intervals}
    \label{fig:latent_rollouts}
\end{figure}

\subsection{Adaptive Policy Training}
Policy training is performed using rollouts from the learned dynamics ensemble. At the beginning of each trajectory, we sample an initial condition and target trajectory from the offline data distribution. We then sample one dynamics model, $T_{\theta_i}$, from the ensemble  where $i={1,...,25}$ and generate an autoregressive rollout by sampling through the probabilistic model $o^\text{off}_{t+1} \sim \mathcal{N}(\mu_{\theta_i} (o^\text{off}_t, a_t) , \sigma_{\theta_i} (o^\text{off}_t, a_t))$.

At each timestep, the current observation is used to form the rollout state and compute rewards. The observation converter then converts the new observation from offline domain to the real-time domain for input to the student observer and policy actor. The latent model updates $z_{t-1}$, and the adaptive policy selects the actuator command, $a_t \sim \pi_{\psi}(a_t \mid o_t^{\mathrm{rt}}, g_t, z_{t-1})$. We follow an asymmetric critic setup, and the critic observes original offline observations $(o_t^{\mathrm{off}})$. The sampled dynamics model then predicts the next original state $s_{t+1}$. The reward is computed using the original state, not the converted observation $r_t = R(o^\text{off}_t)$. This keeps the training objective consistent with the dynamics model while only exposing the actor to real-time-style measurements. Sampling models across the ensemble for trajectory roll-outs trains the policy to perform well across all plausible dynamics, and because regions of high model variance produce more spread in the sampled next states, the policy is exposed to more variable rewards in uncertain regions. As a result, we do not include any explicit uncertainty penalty in the reward.

We optimize the policy using Proximal Policy Optimization~(PPO)~\citep{schulman2017proximal}. The policy is trained entirely in the learned model environment, avoiding online exploratory interaction with the tokamak. The reward function used is a combination of profile tracking error, penalty in changes in normalized plasma pressure $(\beta_N)$. For tracking a given profile $p$, such as rotation profile or electron temperature profile, the overall reward function $R$ is given by:
\begin{align*}
& R(o^\text{off}_t) =
-\left\| e_t \right\|_2^2
- k_B \Delta \beta_{N,t}^2
- k_I \left\| I_t \right\|_2^2,\\
\text{where } &  e_t = p_{\mathrm{target}}(t) - p(t),
\qquad
I_t = 0.9 I_{t-1} + e_t  
\end{align*} 

$\Delta \beta_{N,t} = \beta_{N,t} - \tilde{\beta_N}$ and $\tilde{\beta_N}$ is the mean normalized plasma pressure over the flat top duration of the shot. Since this quantity is representative of overall plasma pressure and confinement, ideally, one would aim to control profiles while minimizing impact to this. By discussion with domain experts and tuning, $k_B$=0.1. The leaky integral term is used to prevent steady state errors, we tune $k_I$ = 0.2.

The policy observation contains the present real-time-style profile estimate, present actuator values, target profile information, tracking error terms, and the latent context $z_{t-1}$. The policy controls changes in actuator commands for NBI (neutral beam injection) power in both on-axis and off axis regions, gas puffing, and ECH power. These actuators provide direct and indirect ways to affect profile evolution. NBI power directly affects temperature and momentum at different regions, gas puffing changes particle input and can affect density and rotation, and ECH modifies temperature and can indirectly affect rotation and transport. More details are provided in Appendix \ref{apnd:policy}. Additionally, we compare policy learning with model predictive control in Appendix \ref{apd:mpc}, where we conclude that MPC is only effective with hyper-parameters that are computationally nonviable for deployment.

\section{Results and Analysis}

In this section, we provide numerical results for the proposed framework. In Section~\ref{sec:5.1}, we show that the latent context extracted from the trained dynamics model encodes useful information about plasma regimes, motivating its use for training the control policy. In Section~\ref{sec:5.2}, we report numerical results for plasma-profile tasks in an offline setting, including rotation and electron temperature profile. In Section~\ref{sec:5.3}, we show that training policies with a state converter improves their performance in a test environment that more closely resembles a real tokamak.

\subsection{Dynamics Context Encodes Useful Plasma Information}
\label{sec:5.1}

We first analyze the latent context extracted from the trained dynamics model. Specifically, we visualize the top three PCA components, which explain approximately 60\% of the variance, against various operating phases and plasma scenarios. In Figure~\ref{fig:shot_phase}, we show that the latent context can well separate two operating phases of fusion plasma: ramp-up and flat-top. 
During ramp-up, the plasma current gradually increases toward its target value, while plasma shape, heating, stored energy evolve toward their desired states. During flat-top, the plasma current is maintained approximately constant, and the plasma operates under relatively stable conditions. These differences in plasma evolution help explain why the learned latent context separates the two phases.

Next, we visualize the latent context across different plasma scenarios in Figure~\ref{fig:confinement_mode}. A plasma scenario refers to a specific operating condition within the tokamak. Scenario labelling has been previously shown by using high frequency (1MHz) BES data \cite{gill2024real}. Here, we see if our latent space developed from low frequency profile data can capture these labels.
The results show that the latent context varies across transitions within the shot from L-mode (Low confinement mode) to H-mode (High confinement mode), then to QH mode (Quiescent H-mode), and finally reaching the WPQH mode (Wide-Pedestal QH mode). The QH mode is characterized as an ELM-free (Edge Localized Mode instabilities) high confinement regime, whereas the WPQH mode maintains a large pedestal width. We notice these variations are much less separated than the shot phase variation (ramp-up vs flat-top), since these operating conditions change quickly over smaller timesteps. e.g. the H mode $\rightarrow$ QH mode $\rightarrow$ WPQH mode $\rightarrow$ H mode transitions can occur within the flat-top phase of a single experiment. Hence, latent space separation is less defined.  In summary, these plots show that the observer latent variation corresponds to physically meaningful plasma changes that influence the dynamics.


\subsection{Dynamics Context Improves Policy Learning}
\label{sec:5.2}


We now investigate whether adding this context in the policy learning stage improves performance. We compare our proposed method against the following baselines: 1) Feedforward, which conditions on the single-timestep observation; 2) History-10, which conditions on the observation-action pairs in the past ten steps; 3) Recurrent GRU, which conditions on the hidden state processed through a recurrent architecture; 4) Time-as-context, which is a feedforward policy with time as the additional context. Policies are trained separately for tracking tasks on two profiles, including rotation and electron temperature.

We compare algorithms in terms of the return (total reward over the trajectory) achieved. The testing environment differs from the training environment in the way sampling is performed from the ensemble of dynamics models. At each timestep during testing, we sample from all members of the ensemble and use the mean as the observation for the policies. Empirically, this strategy gives us the most accurate prediction on the held-out test set.  A possible concern is that our training and testing may exploit the training bias in RPNN (while sampling procedure limits it). To address this we separately create a Transformer Probabilistic Neural Network~(TPNN)-based environment and also use for testing. We test across 43 test shots with 30 repeats each. 

Table~\ref{tab:profile_tracking_rewards} shows the policy performance across these profile tracking tasks on the RPNN-based simulated and TPNN based tokamak environments.
The history-stacked policy does not improve over the feedforward baseline. The recurrent GRU improves over feedforward policy, but does not match the latent-conditioned policy. The time-as-context baseline performs strongly, indicating that shot progression provides useful temporal information. While testing TPNN, we observe that feedforward policy perform better than others (apart from latent conditioned policy). Across the reported mean returns, the latent-conditioned policy performs best on both tasks, across both testing environments.

\begin{table}[!h]
    \centering
    \caption{Performance on the RPNN-based and TPNN-based simulated tokamak environments. Return values are shown, higher (less negative) is better. Standard error in parenthesis.}
    \label{tab:profile_tracking_rewards}
    \resizebox{\linewidth}{!}{
    \begin{tabular}{lcccc}
        \toprule
        & \multicolumn{2}{c}{\textbf{RPNN}}
        & \multicolumn{2}{c}{\textbf{TPNN}} \\
        \cmidrule(lr){2-3} \cmidrule(lr){4-5}
        \textbf{Policy}
            & \textbf{Rotation} & \textbf{Electron Temp}
            & \textbf{Rotation} & \textbf{Electron Temp} \\
        \midrule
        Feedforward
            & $-383.2\,(14.5)$
            & $-183.8\,(13.7)$
            & $-371.5\,(13.1)$
            & $-265.1\,(16.3)$ \\
        History-10 (states + actuators)
            & $-375.1\,(23.1)$
            & $-234.6\,(31.3)$
            & $-468.9\,(14.6)$
            & $-342.0\,(18.2)$ \\
        Recurrent GRU
            & $-361.8\,(35.0)$
            & $-169.0\,(13.1)$
            & $-441.0\,(25.6)$
            & $-308.4\,(19.8)$ \\
        Time-as-context
            & $-315.6\,(17.4)$
            & $-176.5\,(10.4)$
            & $-418.0\,(12.2)$
            & $-271.0\,(16.7)$ \\
        \textbf{Latent-conditioned (ours)}
            & $\mathbf{-298.5\,(17.6)}$
            & $\mathbf{-148.6\,(10.9)}$
            & $\mathbf{-344.6\,(14.2)}$
            & $\mathbf{-249.9\,(16.0)}$ \\
        \bottomrule
    \end{tabular}
    }
\end{table}


To further examine the impact of context conditioning, we roll out several shots using our model for policy tracking. Figure~\ref{fig:latent_rollouts} shows the tracking performance across variations in the latent vector. In each plot, we show the first component of the tracked profile.
For the rotation case in Figure~\ref{fig:rot_variation}, we observe large latent variation at the beginning of the shot. This aligns with the differences between the ramp-up and flat-top plasma phases shown in Figure~\ref{fig:shot_phase}. As a result, the no-latent policy performs poorly at the start, recovers briefly, and then diverges again.
For the electron temperature profile case in Figure~\ref{fig:temp_variation}, the impact at startup is smaller; however, changes in the latent space still correspond to inaccurate tracking. Thus, no-latent policy learns behavior that works reasonably well on average across a wide range of plasma-dynamics regimes. However, latent conditioning allows the policy to adapt to specialized dynamics regions, leading to improved tracking in those regimes.



\subsection{Converter Bridges Real-Time Observation Gap} \label{sec:5.3}

In our experiments so far, we only use offline reconstructed signals (ZIPFIT) for dynamics modelling. We now measure performance in an environment that provides observations from the converter model, which produces domain-shifted signals akin to real-time deployment. We follow the testing environment setup as defined above and test over RPNN and TPNN dynamics.

Table~\ref{tab:converter_impact} shows the performance drop when policies are evaluated with real-time-style observations. Training these policies with the converter setup, maintains performance in a real-time like environment.

\begin{table}[t]
\centering
\caption{Impact of converter training on RT-style profile tracking for the
latent-conditioned policy under two dynamics backbones. Evaluation return
(higher is better) with standard errors in parentheses.}
\label{tab:converter_impact}
\small
\setlength{\tabcolsep}{5pt}
\begin{adjustbox}{max width=\linewidth}
\begin{tabular}{@{}lrrrr@{}}
\toprule
 & \multicolumn{2}{c}{\textbf{RPNN Test Env}} & \multicolumn{2}{c}{\textbf{TPNN Test Env}} \\
\cmidrule(lr){2-3}\cmidrule(l){4-5}
\textbf{RT Tracking Task} & Trained w/o converter & Trained w converter
                 & Trained w/o converter & Trained w converter \\
\midrule
Rotation    & $-1032.1$ (29.3) & $\mathbf{-405.1}$ (14.8) & $-1544.4$ (29.5) & $\mathbf{-482.4}$ (13.6) \\
Electron Temp & $-353.4$ (16.4)  & $\mathbf{-246.9}$ (9.1)  & $-603.5$ (15.4)  & $\mathbf{-357.0}$ (12.3) \\
\bottomrule
\end{tabular}
\end{adjustbox}
\end{table}

\section{Experiment Results at DIII-D Tokamak}
We finally tested our algorithm on the DIII-D tokamak in two experimental shots. Due to operation limitations, we only report results from the final latent-conditioned policy with the observation converter. Quantitative results are summarized in Table \ref{tab:whole_window_tracking_metrics}. These experiments serve as demonstrations rather than statistical significant improvement. 

\begin{table}[!h]
    \centering
    \caption{Quantitative tracking metrics over real world results. Metrics computed over controller window 1500--4000 ms.}
    \label{tab:whole_window_tracking_metrics}
    \begin{tabular}{lcc}
        \toprule
        \textbf{Metric} & \textbf{Rotation (206591)} & \textbf{Te (206589)} \\
        \midrule
        RMSE             & 27.41 km/s  & 0.722 keV  \\
        MAE              & 18.83 km/s  & 0.579 keV  \\
        Mean abs.\ target & 139.49 km/s & 1.395 keV  \\
        Mean bias        & +10.09 km/s & +0.561 keV \\
        NRMSE            & 0.196 & 0.518 \\
        \bottomrule
    \end{tabular}
\end{table}


\begin{figure}[t]
    \centering
    \begin{subfigure}[t]{0.44\linewidth}
        \centering
        \includegraphics[width=\linewidth]{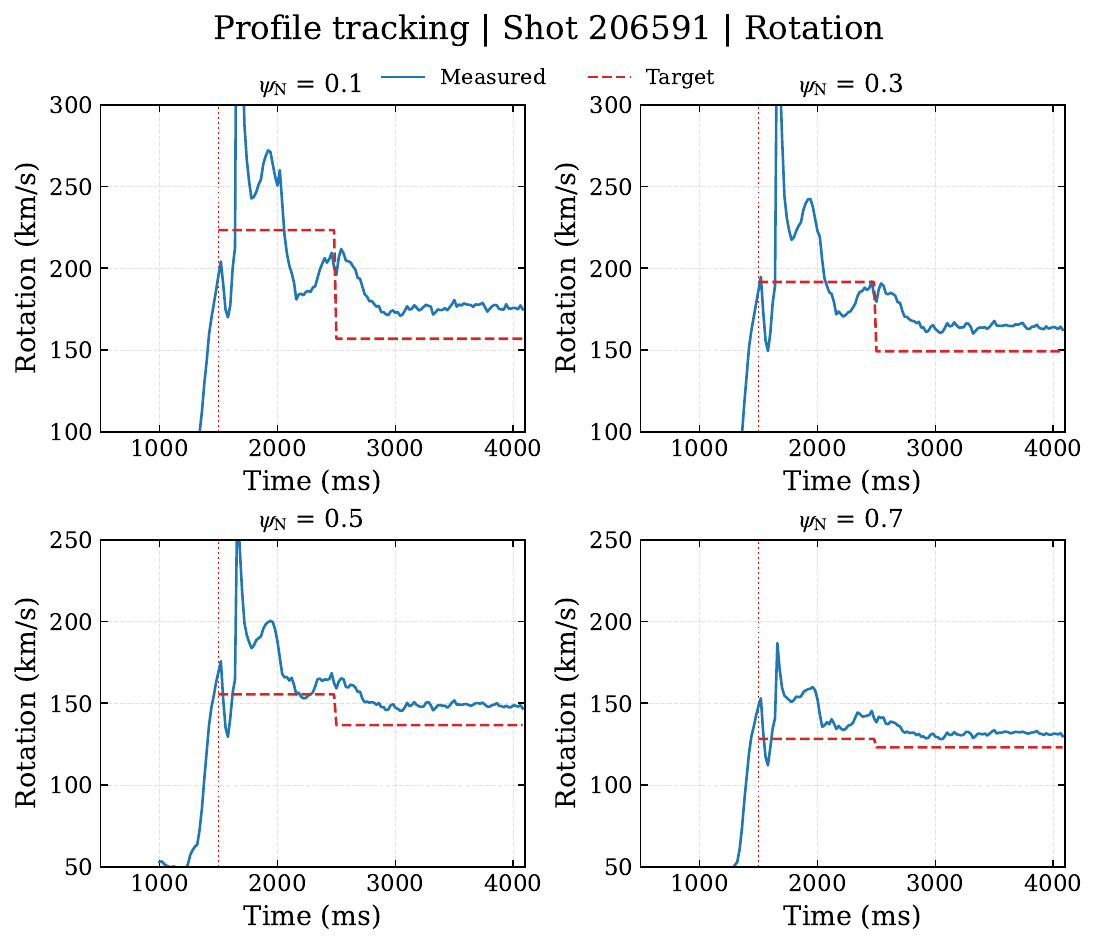}
        \caption{Rotation-profile tracking at various normalized flux radius $(\Psi_N)$ (from core to outer edge of plasma).}
        \label{fig:rot_variation_s}
    \end{subfigure}
    \qquad
    \begin{subfigure}[t]{0.44\linewidth}
        \centering
        \includegraphics[width=\linewidth]{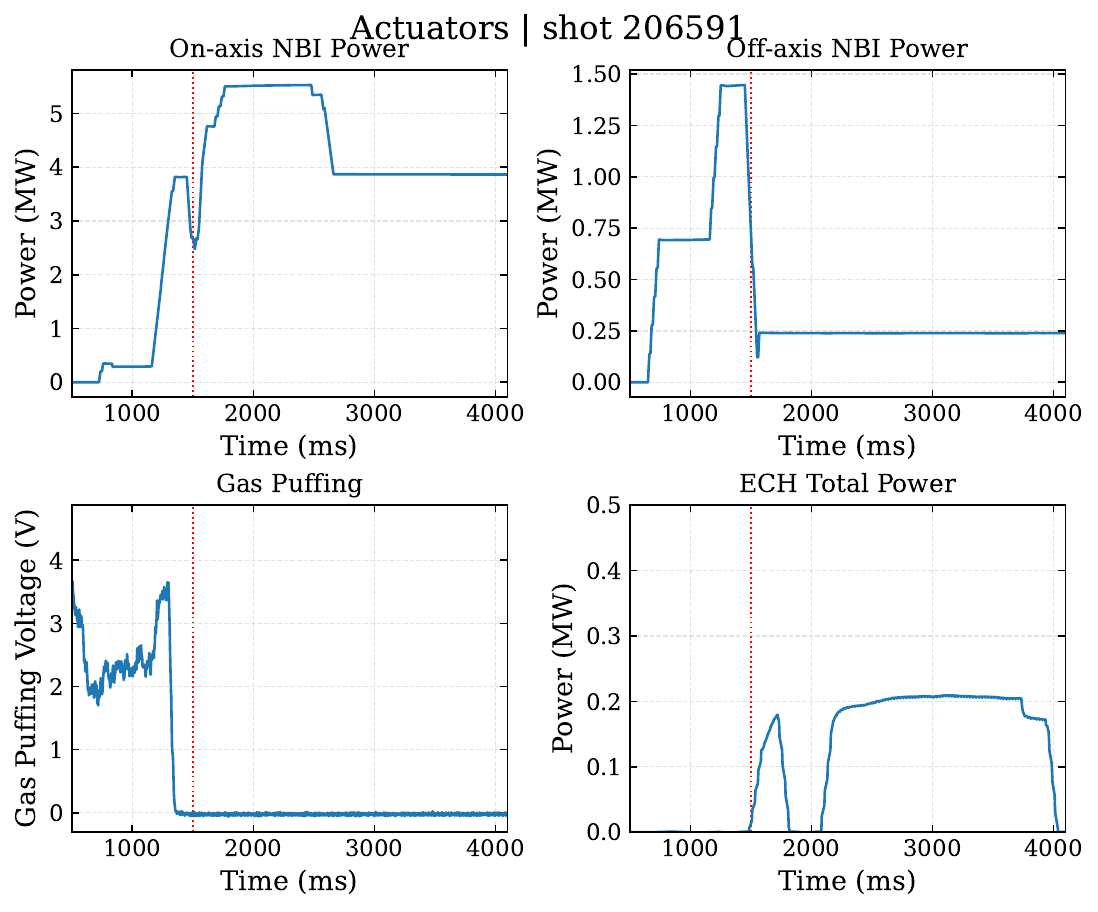}
        \caption{Actuator Response}
        \label{fig:rot_variation_a}
    \end{subfigure}

    \caption{Experimental result for rotation profile-tracking at DIII-D Tokamak: Shot 206591}
    \label{fig:diid-d-rot}
\end{figure}

\begin{figure}[t]
    \centering
    \begin{subfigure}[t]{0.44\linewidth}
        \centering
        \includegraphics[width=\linewidth]{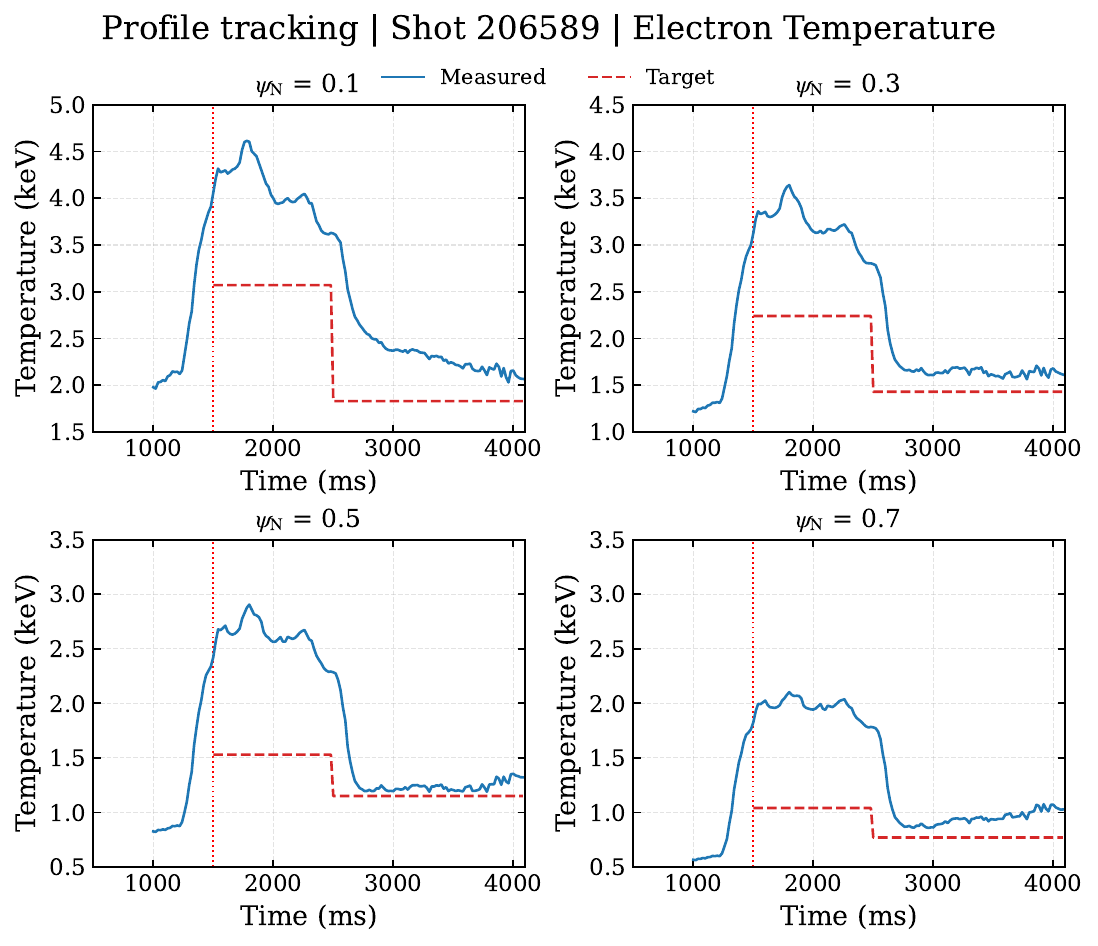}
        \caption{Electron temperature-profile tracking at various normalized flux radius $(\Psi_N)$(from core to outer edge of plasma).}
        \label{fig:temp_variation_s}
    \end{subfigure}
    \qquad
    \begin{subfigure}[t]{0.44\linewidth}
        \centering
        \includegraphics[width=\linewidth]{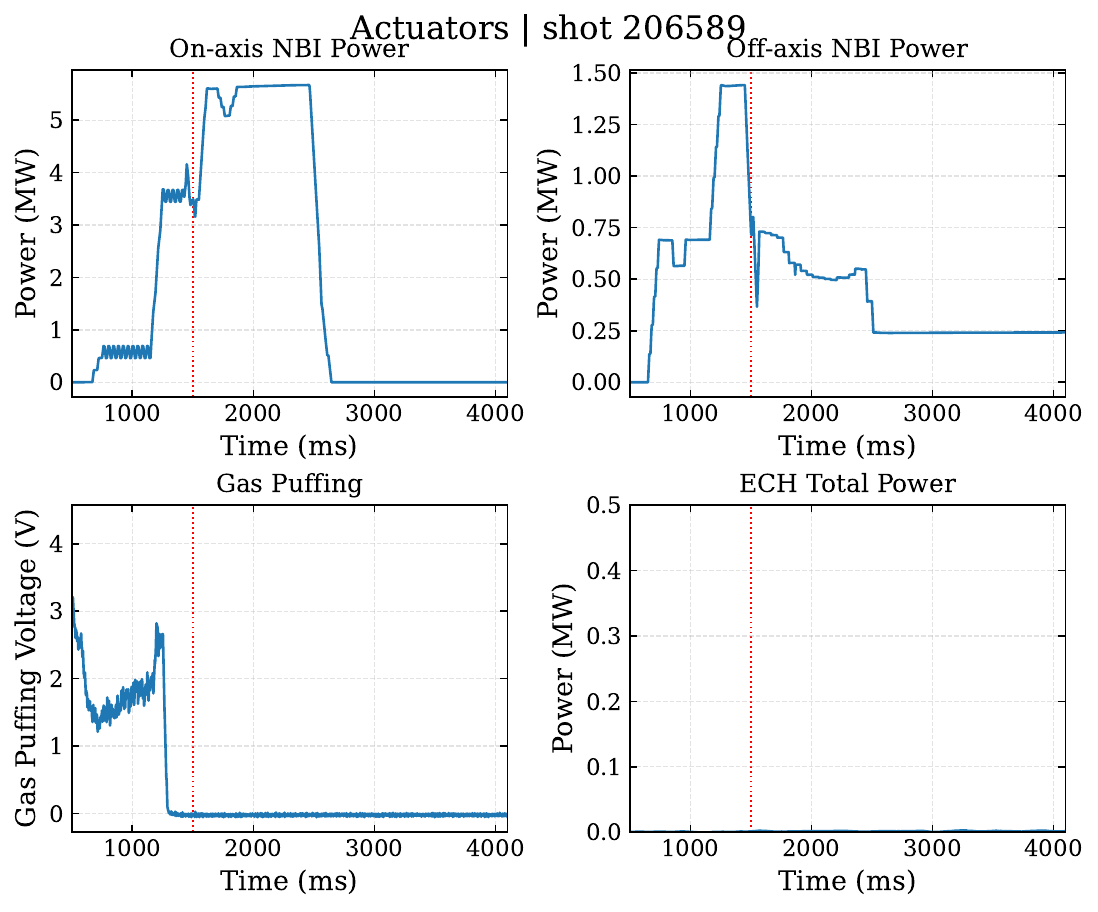}
        \caption{Actuator Response}
        \label{fig:temp_variation_a}
    \end{subfigure}

    \caption{Experimental result for electron temperature profile-tracking at the DIII-D Tokamak: Shot 206589}
    \label{fig:diii-d-temp}
\end{figure}

\textbf{Shot 206591: Rotation profile tracking (Figure~\ref{fig:diid-d-rot}):}\quad NBI (neutral beam injection) delivers power in the toroidal direction, causing the plasma to rotate. Thus, on and off-axis NBI have strongest effects on controlling rotation. We track a single-step target change in the experiment. Our controller starts acting on the actuators at 1500ms. Initially a large spike in rotation is recorded - this spike is largely attributed to diagnostic errors. The policy action attempts to bring it down by reducing off axis NBI, and overshoots to reduce it (2200ms). It then uses ECH to bring the rotation back up. We hypothesize the controller learns that increasing ECH power leads to reduced density, and therefore reduced inertia, thus leading to higher rotation. Once the target shifts (2500ms), on-axis power decreases and ECH is maintained to stabilize the rotation and prevent betaN from decreasing too much. Note objective can be achieved in multiple ways, however, this is multi-input multi-output control (MIMO) control setting and the policy learns one solution out of many. 


\textbf{Shot 206589: Electron Temperature ($T_e$) profile tracking~(Figure~\ref{fig:diii-d-temp}):}\quad On and Off-axis NBI deposit energy to the plasma, increasing the electron temperature. ECH directly heats the electrons, thus increasing electron temperature. Our controller starts at 1500ms, however, this is a more challenging setting since the $T_e$ is already on a sharp increase. Immediately off-axis NBI is reduced.
When the target shifts at 2500ms we see sharp changes in both on-axis and off-axis NBI which brings $T_e$ down. No ECH is used because the profile remains above the target; therefore, additional heating is not required.  

\section{Conclusion, Limitations and Future Work}







We propose an offline RL framework for real-time tokamak profile control that addresses partial observability and diagnostic mismatch. The policy is conditioned on a dynamics-informed latent context from an RPNN plasma model, enabling adaptation across plasma regimes. An observation converter further aligns policy training with real-time diagnostics. Numerical results show improved profile-tracking performance over no-latent policies, and DIII-D experiments provide initial real-time validation. 
However, we acknowledge the following limitations - 
\begin{itemize}[leftmargin=*]
    \item \textbf{Limited Real World Testing:} Two shots demonstrate execution on the real PCS using real-time diagnostics, but broader testing is needed to establish performance statistically. DIII-D access is competitive and rare, which also motivates our direction of developing accurate surrogates and adaptable policies which support out-of-the-box deployment.

    \item \textbf{Actuator and Sensor Dropouts:} These failures are common in real world testing but are not yet modeled during training. Explicitly modelling them would lead to more robust policies.

    \item \textbf{Reliance on RTCAKENN:} Deployment currently requires RTCAKENN and its diagnostic inputs. Modeling raw diagnostics directly could further simplify policy learning and deployment.
\end{itemize}

Addressing these limitations by introducing more sensor and actuator dropouts in training leading to robust control are important directions of future work. Another direction is to explore potential applications of the learned latent context to related tasks, such as tearing-instability prediction or regime-transition detection.


\section*{Acknowledgement}
This work would not have been possible without support from many people. We wish to thank the incredible staff at DIII-D and Princeton Plasma Physics Lab (PPPL). We also thank Keith Erickson, Plasma Control System Team, Azarakhsh Jalalvand and SangKyeun Kim for their help during deployment. This material is based upon work supported by the U.S. Department of Energy, Office of Science, Office of Fusion Energy Sciences, using the DIII-D National Fusion Facility, a DOE Office of Science user facility, under Awards DE-FC02-04ER54698, DE-SC0024544 and DE-SC0015480. Rohit was also supported by the Indian Institute of Technology Alumni fellowship.

\section*{Disclaimer}
This report was prepared as an account of work sponsored
by an agency of the United States Government. Neither
the United States Government nor any agency thereof, nor
any of their employees, makes any warranty, express or implied, or assumes any legal liability or responsibility for the
accuracy, completeness, or usefulness of any information,
apparatus, product, or process disclosed, or represents that
its use would not infringe privately owned rights. Reference
herein to any specific commercial product, process, or service by trade name, trademark, manufacturer, or otherwise
does not necessarily constitute or imply its endorsement, recommendation, or favoring by the United States Government
or any agency thereof. The views and opinions of authors
expressed herein do not necessarily state or reflect those of
the United States Government or any agency thereof.

\bibliographystyle{apalike}
\bibliography{ref}

\appendix

\section{Dynamics Model Training}\label{apnd:rpnn}

\subsection{State and Actuator Space -}
The dynamics model (an ensemble of RPNN networks) models the transitions on a larger state and action space, in comparison to the policy. This is shown in table \ref{tab:signals}. Variables marked under Actuators and States are used in dynamics modeling. Only a subset of the states is provided as observations to the policy (marked under observations), and a subset of actuators is controlled by the policy (marked under actions).  We do this to simplify the learning process and also because other actuators have different purposes other than controlling the selected task. Note that last action taken is provided in the observation space of the policy.

Spatial profile quantities are reduced using PCA before being used in dynamics models, policy observations, and target representations. Following prior work, 33D profiles are projected to a small number of PCA coefficients. For the main profile quantities, four PCA components are used for electron temperature, ion temperature, density, and rotation. Pressure and safety-factor profiles use two components where applicable. These reduced representations capture more than 99\% of the variance in the corresponding offline profile data.

\begin{table}[t]
\centering
\caption{Plasma signals and how they are used as State and Actuators for dynamics modelling. Policy Observation and action space variables are also shown. Note that rotation and electron temperature are only observable to the policies that control them, not generally.}
\label{tab:signals}
\footnotesize
\setlength{\tabcolsep}{3pt}
\renewcommand{\arraystretch}{1.1}
\begin{tabular}{@{}>{\raggedright\arraybackslash}p{0.16\linewidth} >{\raggedright\arraybackslash}p{0.36\linewidth} cccc@{}}
\toprule
\textbf{Signal Group} & \textbf{Signals} &
\textbf{Actuator} & \textbf{State} &
\textbf{Action} & \textbf{Observation} \\
\midrule
Scalar States
  & $\beta_{N}$ (normalized plasma pressure), $l_i$ (internal inductance),
    line-averaged density, loop voltage, MHD stored energy
  & \xmark & \cmark & \xmark & \cmark \\
\midrule
Profile States
  & Rotation, electron temperature
  & \xmark & \cmark & \xmark & \cmark \\
  & Density, ion temperature, pressure, safety factor $q$
  & \xmark & \cmark & \xmark & \xmark \\
\midrule
Shape Variables
  & Elongation, upper and bottom triangularity, $a_{\text{minor}}$,
    radial and vertical positions of magnetic axis
  & \cmark & \xmark & \xmark & \xmark \\
\midrule
Neutral Beam Variables
  & On-axis power, off-axis power
  & \cmark & \xmark & \cmark & \cmark \\
\midrule
Gas Puffing
  & GasA voltage
  & \cmark & \xmark & \cmark & \cmark \\
\midrule
Electron Cyclotron Heating
  & ECH total power
  & \cmark & \xmark & \cmark & \cmark \\
\midrule
Other Actuators
  & Current target, toroidal field
  & \cmark & \xmark & \xmark & \xmark \\
\midrule
Targets
  & Rotation target ($t$), rotation target ($t+10$), current error terms,
    leaky integral error
  & \xmark & \xmark & \xmark & \cmark \\
\bottomrule
\end{tabular}
\end{table}

\subsection{Recurrent Probabilistic Neural Network (RPNN) }

\textbf{Network Architecture : }
\begin{itemize}
    \item \textbf{Encoder:}
    \begin{itemize}
        \item Fully Connected (FC) layer: \texttt{input\_dim} $\times$ 512
        \item FC layer: 512 $\times$ 512
    \end{itemize}
    
    \item \textbf{Memory Unit:}
    \begin{itemize}
        \item Gated Recurrent Unit (GRU) block: 512 $\times$ 256
    \end{itemize}

    \item \textbf{Decoder (with residual connections between FC layers):}
    \begin{itemize}
        \item FC layer: 256 $\times$ 512
        \item FC layers: 512 $\times$ 512 (repeated 8 times)
        \item FC layer: 512 $\times$ 128
    \end{itemize}

    \item \textbf{Output Heads:}
    \begin{itemize}
        \item Mean head: 128 $\times$ \texttt{output\_dim}
        \item Log-variance head: 128 $\times$ \texttt{output\_dim}
    \end{itemize}
\end{itemize}

The network predicts the parameters of a probability distribution, and is trained in two stages. First stage corresponds to training with Mean Squared error loss (MSE) and second stage corresponds to training only the log variance head with negative log likelihood loss. We use the Adam optimizer with a learning rate of $3 \times 10^{-4}$ and a weight decay of $1 \times 10^{-3}$ in both stages. A 25 model ensemble is trained by bootstrapping on the dataset. We train for 1000 epochs with early stopping (patience = 250 epochs) based on performance on a validation set comprising 5\% of the total data. Each RPNN model took ~8 h on an NVIDIA A6000. The 25-member ensemble was parallelized across 7 A6000 GPUs and completed in ~8 hours.

\subsection{Transformer Probabilistic Neural network (TPNN)}

We create a dynamics model with a different backbone to rule out possibilities of policy learning to exploit the RPNN model bias. Hence we create a transformer probabilistic Neural network (TPNN).

\textbf{Network Architecture : }
\begin{itemize}
    \item \textbf{Encoder:}
    \begin{itemize}
        \item Fully Connected (FC) layer: \texttt{input\_dim} $\times$ 512
        \item FC layer: 512 $\times$ 512
        \item Layer normalization, followed by a learned positional encoding
              over a 225-step context
    \end{itemize}

    \item \textbf{Memory Unit:}
    \begin{itemize}
        \item GPT-style causal self-attention blocks (repeated 2 times), each
              with model width 512 and 8 attention heads (64 dimensions per head)
        \item Each block applies pre-layer normalization, multi-head causal
              self-attention, and a feed-forward network of
              512 $\times$ 2048 and 2048 $\times$ 512, with residual
              connections and dropout of 0.2
    \end{itemize}

    \item \textbf{Decoder (with residual connections between blocks):}
    \begin{itemize}
        \item FC block: 512 $\times$ 512 $\times$ 512
        \item Residual FC blocks: 512 $\times$ 512 $\times$ 512 (repeated 3 times)
        \item FC block: 512 $\times$ 512 $\times$ 128
    \end{itemize}

    \item \textbf{Output Heads:}
    \begin{itemize}
        \item Mean head: 128 $\times$ \texttt{output\_dim}
        \item Log-variance head: 128 $\times$ \texttt{output\_dim}
    \end{itemize}
\end{itemize}

The TPNN replaces the recurrent
memory unit of the RPNN with self-attention over a buffer of the past 225 inputs,
and keeps the same inputs, outputs, and data split. Like the RPNN, it predicts
the parameters of a probability distribution and is trained in two stages. The
first stage trains with Mean Square Error (MSE) loss, and the second stage
freezes all other weights and trains only the log-variance head with negative
log likelihood loss. We use the AdamW optimizer with a learning rate of
$3 \times 10^{-4}$ and a weight decay of $1 \times 10^{-3}$ in both stages, with
an effective batch size of 1024 sequences of 225 steps. We train each stage for up to 1000 epochs with early stopping
(patience = 150 epochs in the first stage and 100 epochs in the second) based
on performance on a validation set comprising 5\% of the total data. The first
stage took $\sim$2.9 h and the second stage took $\sim$5.5 h on a single NVIDIA A6000.

\subsection{RPNN \& TPNN Performance}

We report explained variance numbers over the next state prediction. These numbers are over a test set of 10\% of shots.
These are shown in Table \ref{tab:dynamics_fullstate_ev_all}. We note the RPNN outperforms the TPNN and hence it is our main test environment. The TPNN serves as a secondary model to check against overfitting the model bias. We note that all major profile components are predicted well, hence our setup is reliable for autoregressive rollouts of shot trajectories. More in depth analysis of RPNN is presented in \citep{char2024full}. 

\begin{table}[t]
\centering
\caption{Open-loop full-state explained variance for all 27 state variables at
rollout horizons of 1, 5, 10, and 20 steps. For profiles, c1--c4 are PCA
components, and parentheses denote the profile variance represented by each
component. Both RPNN and TPNN are
evaluated on the same held-out trajectories. The last row is the mean over the
27 states.}
\label{tab:dynamics_fullstate_ev_all}
\small
\setlength{\tabcolsep}{5pt}
\begin{adjustbox}{max width=\linewidth, max totalheight=0.9\textheight}
\begin{tabular}{@{}lrrrrrrrr@{}}
\toprule
 & \multicolumn{4}{c}{RPNN} & \multicolumn{4}{c}{TPNN} \\
\cmidrule(lr){2-5}\cmidrule(l){6-9}
State (variance) & h1 & h5 & h10 & h20 & h1 & h5 & h10 & h20 \\
\midrule
\multicolumn{9}{@{}l}{\textit{Electron temperature ($T_e$)}} \\
\quad c1 (95.06\%) & .995 & .949 & .890 & .862 & .993 & .931 & .842 & .842 \\
\quad c2 (4.19\%)  & .963 & .771 & .665 & .518 & .955 & .660 & .544 & .370 \\
\quad c3 (0.63\%)  & .940 & .683 & .598 & .517 & .925 & .518 & .486 & .408 \\
\quad c4 (0.31\%)  & .902 & .588 & .469 & .092 & .884 & .403 & .369 & $-.182$ \\
\addlinespace
\multicolumn{9}{@{}l}{\textit{Ion temperature ($T_i$)}} \\
\quad c1 (96.34\%) & .974 & .934 & .889 & .852 & .973 & .933 & .897 & .789 \\
\quad c2 (2.88\%)  & .968 & .829 & .792 & .580 & .963 & .830 & .770 & .432 \\
\quad c3 (0.88\%)  & .883 & .634 & .706 & .721 & .860 & .612 & .685 & .639 \\
\quad c4 (0.23\%)  & .537 & .064 & .206 & .242 & .491 & .133 & .213 & .054 \\
\addlinespace
\multicolumn{9}{@{}l}{\textit{Electron density ($n_e$)}} \\
\quad c1 (94.41\%) & .999 & .992 & .980 & .943 & .999 & .991 & .969 & .886 \\
\quad c2 (4.13\%)  & .984 & .747 & .557 & .262 & .980 & .772 & .472 & .244 \\
\quad c3 (1.25\%)  & .915 & .706 & .483 & .210 & .923 & .629 & .260 & $-.067$ \\
\quad c4 (0.52\%)  & .917 & .627 & .424 & .225 & .939 & .703 & .406 & .074 \\
\addlinespace
\multicolumn{9}{@{}l}{\textit{Toroidal rotation}} \\
\quad c1 (92.31\%) & .991 & .900 & .850 & .648 & .991 & .909 & .846 & .617 \\
\quad c2 (5.52\%)  & .959 & .766 & .744 & .473 & .957 & .726 & .655 & .290 \\
\quad c3 (1.50\%)  & .784 & .493 & .460 & .492 & .808 & .427 & .431 & .294 \\
\quad c4 (0.65\%)  & .699 & .293 & .304 & .187 & .692 & .217 & .320 & $-.492$ \\
\addlinespace
\multicolumn{9}{@{}l}{\textit{Pressure ($p$)}} \\
\quad c1 (97.29\%) & .999 & .991 & .982 & .969 & .999 & .987 & .955 & .832 \\
\quad c2 (2.71\%)  & .983 & .898 & .781 & .605 & .979 & .874 & .551 & .341 \\
\addlinespace
\multicolumn{9}{@{}l}{\textit{Safety factor ($q$)}} \\
\quad c1 (97.02\%) & .997 & .978 & .963 & .936 & .997 & .975 & .958 & .881 \\
\quad c2 (2.98\%)  & .993 & .941 & .848 & .699 & .991 & .938 & .859 & .680 \\
\midrule
Normalized beta ($\beta_N$)           & .998  & .987 & .971 & .945 & .998  & .983 & .939 & .854 \\
Line-averaged density ($\bar{n}_e$)   & 1.000 & .992 & .980 & .943 & 1.000 & .990 & .966 & .874 \\
Internal inductance ($l_i$)           & .997  & .981 & .917 & .575 & .997  & .982 & .930 & .624 \\
Central safety factor ($q_0$)         & .998  & .976 & .964 & .941 & .995  & .967 & .958 & .924 \\
Edge safety factor ($q_{95}$)         & 1.000 & .999 & .998 & .994 & 1.000 & .998 & .996 & .987 \\
Loop voltage ($V_{\mathrm{loop}}$)    & .936  & .880 & .844 & .699 & .931  & .806 & .771 & .704 \\
Stored energy ($W_{\mathrm{MHD}}$)    & .998  & .987 & .972 & .940 & .998  & .981 & .935 & .816 \\
\midrule
Mean & .937 & .799 & .749 & .632 & .934 & .773 & .703 & .563 \\
\bottomrule
\end{tabular}
\end{adjustbox}
\end{table}

\section{Policy Learning}\label{apnd:policy}

Policy training is performed in a learned model environment rather than on the tokamak. Each synthetic trajectory starts from an initial condition sampled from historical data. A dynamics model from the RPNN ensemble is sampled and rolled forward autoregressively. At every timestep, the policy selects an action, the learned dynamics model predicts the next reconstructed state, and the reward is computed from the reconstructed state. The latent vector is lagged for strict causality. It is initialized with a zero vector.

 The final latent-conditioned actor receives the 33-dimensional real-time policy observation together with a learned 256-to-32 dimension projection of the observer latent, giving a 65-dimensional actor input. The actor uses two 500-unit hidden layers and outputs four actuator commands. The critic network is similar sized but with single value output. Training the observer and converter takes ~1 hour on an A6000, and training our latent-conditioned policy takes ~22 hours on one A6000.

\begin{algorithm}[h]
\caption{Adaptive Offline Model-Based RL Training. We use a lagged latent vector to maintain causality (initial values are zeros).}
\begin{algorithmic}[1]
\State Train RPNN dynamics ensemble on historical reconstructed DIII-D data.
\State Train observation converter $C_\psi: O^{off}\rightarrow O^{rt}$.
\State Train student latent model $S_\phi$ to predict teacher dynamics hidden states.
\For{each policy update}
    \State Sample initial reconstructed state from historical data.
    \State Sample target trajectory $g_{0:H}$.
    \State Sample one RPNN dynamics model from the ensemble.
    \For{$t=0,\ldots,H-1$}
        \State Convert reconstructed observation: $\hat{o}^{rt}_t = C_\psi(o^{off}_t)$.
        \State Update latent context: $z_{t-1} = S_\phi(\hat{o}^{rt}_{t-1}, a_{t-1}, z_{t-2})$.
        \State Sample action: $a_t \sim \pi_\theta(\hat{o}^{rt}_t, g_t, z_{t-1})$.
        \State Predict next reconstructed state using RPNN dynamics.
        \State Compute reward using reconstructed state and target.
    \EndFor
    \State Update policy using PPO.
\EndFor
\end{algorithmic}
\end{algorithm}

The policy observation contains the present real-time-style profile estimate, present actuator values, target profile information, tracking error terms, and the latent context $z_t$. In the rotation-profile setting, the policy observed the RTCAKENN rotation profile estimate, current NBI on-axis and off-axis power, GasA voltage, ECH total power, present target, future target shifted 10 timesteps ahead, and current tracking error. In the adaptive profile-control setting, the same structure is extended to profile quantities such as rotation, temperature, and density.

The policy controls changes in actuator commands for neutral beam on axis and off axis power, gas puffing, and ECH power. These actuators provide direct and indirect ways to affect profile evolution. Neutral beam power directly affects heating and momentum input at different locations, gas puffing changes particle input and can affect density and rotation, and ECH modifies temperature and can indirectly affect rotation and transport.

To avoid unrealistic targets, target profiles are sampled from historical discharges with similar operating conditions. In the rotation-profile setting, a reference shot is replayed and another shot from the same experimental session is selected as the target shot. Two time points from the target shot are used to define a step-function target. This ensures the sampled targets remain physically plausible for the reference scenario while still producing diverse tracking tasks

\subsection{Student Observer Model}\label{apd:observer}

Table \ref{table:student-latent-model} shows the model details for the student observer model. We train the model to predict the mean of the ensemble latent. This choice is motivated by its role as an empirically stable learning target. To test this directly, we replaced the averaged target with a single ensemble member's hidden state which is sampled randomly when generating a sequence during policy learning. Table \ref{tab:observer_target} shows that sampling individual ensemble members increases target variance by 38× and reduces student reconstruction EV from 0.643 to 0.009. Thus, we use averaging as an empirically validated aggregation choice rather than assuming alignment between individual ensemble representations.

\begin{table}[H]
\centering
\caption{Student latent model / hidden-state predictor training details.}
\label{table:student-latent-model}
\small
\begin{tabular}{ll}
\toprule
Item & Value \\
\midrule
Model & GRU hidden-state predictor \\
Target & Teacher RPNN ensemble-mean hidden state \\
Input & State, current actuator, next actuator \\
Loss & Mean squared error \\
Default encoder dimension & 128 \\
Default GRU hidden size & 32 \\
Default GRU layers & 2 \\
Default batch size & 32 \\
Default sequence length & 50 \\
Default learning rate & $10^{-3}$ \\
Early stopping & Validation loss based \\
\bottomrule
\end{tabular}
\end{table}

\begin{table}[h]
\centering
\caption{Student observer trained against the elementwise mean
of the ensemble hidden states versus a randomly sampled member per sequence. Reconstruction explained variance (EV) and target variance are measured on a held out set.}
\label{tab:observer_target}
\small
\begin{tabular}{@{}lrr@{}}
\toprule
Teacher target & Reconstruction EV & Target variance \\
\midrule
Fixed elementwise mean  & 0.643 & 0.0086 \\
Random member/sequence  & 0.009 & 0.328 \\
\bottomrule
\end{tabular}
\end{table}

\subsection{Observation Converter}\label{apd:converter}

The converter addresses the diagnostic-domain shift between offline reconstructed profiles and real-time diagnostic estimates. Dynamics and reward computation use offline reconstructed states because they are smoother and more reliable for long-horizon model learning. The policy, however, must operate on real-time observations at deployment. The converter maps reconstructed observations to RTCAKENN-style observations during training so the policy sees inputs closer to those available on the tokamak.

\begin{table}[h]
\centering
\caption{Input and output features for the simple observation converter.}
\small
\begin{tabular}{lll}
\toprule
Group & Offline input features & Real-time output features \\
\midrule
Density & dens\_component1--4 & dens\_RTCAKENN\_component1--4 \\
Electron temperature & temp\_component1--4 & temp\_RTCAKENN\_component1--4 \\
Ion temperature & itemp\_component1--4 & itemp\_RTCAKENN\_component1--4 \\
Rotation & rotation\_component1--4 & rot\_RTCAKENN\_component1--4 \\
Pressure & pres\_EFIT01\_component1--2 & pres\_RTCAKENN\_component1--2 \\
\bottomrule
\end{tabular}
\end{table}

To evaluate performance of the converter, we measure the held out set explained variance, along with miscalibration area and, negative log likelihood which estimates the likelihood of real RTCAKENN data under the converter's predictions. We also compare variations of the model where a constant variance is used and not learnt. This is shown in Table \ref{tab:converter_calibration}. The converter performs well, and the heteroscedastic model (our approach) achieves lower NLL than the constant-variance baseline. We sample from this predicted distribution during policy training.

\begin{table}[H]
\centering
\caption{Pooled held-out accuracy and uncertainty calibration of theZIPFIT-to-RTCAKENN converter. EV measures explained variation (higher is better). cov68/cov95 give empirical coverage within predicted $\pm 1\sigma$/$\pm 1.96\sigma$ intervals (nominally 68\%/95\%). Miscalibration
area measures deviation from expected coverage. NLL measures the likelihood of held-out RTCAKENN observations under the converter's predicted distributions
(lower is better).}
\label{tab:converter_calibration}
\small
\begin{tabular}{@{}lr@{}}
\toprule
Metric & Result \\
\midrule
EV                    & 0.655 \\
cov68                 & 0.744 \\
cov95                 & 0.964 \\
Miscalibration area   & 0.040 \\
Heteroscedastic NLL   & 0.637 \\
Constant-$\sigma$ NLL & 0.797 \\
\bottomrule
\end{tabular}
\end{table}

Additionally, we compare policy actions from converter-generated observations with actions from actual recorded RTCAKENN observations at paired historical timesteps. This is teacher-forced because after a counterfactual policy action, the recorded next RTCAKENN observation corresponds to the historical action rather than the policy action. This is shown in Table \ref{tab:converter_action_replay}. The policy actions obtained from converted and actual RTCAKENN observations are strongly correlated. In addition, the action obtained from the actual RTCAKENN observation lies within the action range induced by converter sampling for 83\% of rotation samples and 86\% of electron-temperature samples. Thus, the policy behavior produced through the learned observation converter is generally consistent with the behavior obtained from the recorded real-time reconstruction.

\begin{table}[H]
\centering
\caption{Teacher-forced policy-action replay on recorded real-time data. Action correlation compares the policy's actions under converter observations with its
actions under the recorded RTCAKENN observations. The last column gives the fraction of actual-RT actions that fall within the band of actions produced from converter samples.}
\label{tab:converter_action_replay}
\small
\begin{tabular}{@{}lrr@{}}
\toprule
Task & Action correlation & \shortstack[r]{Actual-RT action within\\converter-sampled band} \\
\midrule
Rotation             & 0.968 & 83\% \\
Electron temperature & 0.968 & 86\% \\
\bottomrule
\end{tabular}
\end{table}

\subsection{Policy Learning Ablations}
While the baseline policies mentioned in Table \ref{tab:profile_tracking_rewards} only observe the target profiles, we also perform ablation study by giving baseline policies data from other profiles as well. This is to compare against our dynamics latent conditioned policy which takes in information from other profiles through student observer model. In feedforward policies, adding data from all profiles causes a decline in performance showing that additional information without recurrent architecture does not lead to improvements. In the recurrent GRU policy we do see a slight improvement, but within error ranges, indicating that temporal context is important for performance. In contrast, the observer module predicts a latent context to match hidden state of the RPNN dynamics model. This hidden state encodes useful information for policy performance, which our method is able to utilize. 

\begin{table}[!h]
    \centering
    \caption{Expanded results on Performance on the RPNN-based and TPNN-based simulated tokamak environments. (full) refers to observing all profiles. Our latent conditioned policy only observes target policy but all profiles are input to observer module which creates the latent context. Return values are shown, higher (less negative) is better. Standard error in parenthesis.}
    \label{tab:profile_tracking_rewards_expanded}
    \resizebox{\linewidth}{!}{
    \begin{tabular}{lcccc}
        \toprule
        & \multicolumn{2}{c}{\textbf{RPNN}}
        & \multicolumn{2}{c}{\textbf{TPNN}} \\
        \cmidrule(lr){2-3} \cmidrule(lr){4-5}
        \textbf{Policy}
            & \textbf{Rotation} & \textbf{Electron Temp}
            & \textbf{Rotation} & \textbf{Electron Temp} \\
        \midrule
        Feedforward
            & $-383.2\,(14.5)$
            & $-183.8\,(13.7)$
            & $-371.5\,(13.1)$
            & $-265.1\,(16.3)$ \\
        Feedforward (full)
            & $-400.5 \,(35.1)$
            & $-247.6\,(32.9)$
            & $-454.1\,(17.8)$
            & $-284.4\,(22.6)$ \\
        History-10 (states + actuators)
            & $-375.1\,(23.1)$
            & $-234.6\,(31.3)$
            & $-468.9\,(14.6)$
            & $-342.0\,(18.2)$ \\
        History-10 (full)
            & $-557\,(23.7)$
            & $-235.9\,(14.6)$
            & $-524.6\,(26.5)$
            & $-475.3\,(40.1)$ \\
        Recurrent GRU
            & $-361.8\,(35.0)$
            & $-169.0\,(13.1)$
            & $-441.0\,(25.6)$
            & $-308.4\,(19.8)$ \\
        Recurrent GRU (full)
            & $-356.8\,(69.0)$
            & $-162.0\,(10.8)$
            & $-430.1\,(22.4)$
            & $-297.2\,(14.0)$ \\
        Time-as-context
            & $-315.6\,(17.4)$
            & $-176.5\,(10.4)$
            & $-418.0\,(12.2)$
            & $-271.0\,(16.7)$ \\
        \textbf{Latent-conditioned (ours)}
            & $\mathbf{-298.5\,(17.6)}$
            & $\mathbf{-148.6\,(10.9)}$
            & $\mathbf{-344.6\,(14.2)}$
            & $\mathbf{-249.9\,(16.0)}$ \\
        \bottomrule
    \end{tabular}
}
\end{table}

\section{Model Predictive Control}\label{apd:mpc}

As an additional baseline we implement a shooting-based MPC using our learned dynamics model. We varied the number of sampled action sequences ($K$) and planning horizon ($H$) using Cross Entropy method (CEM) planning. Planning is performed using a single ensemble member sampled at each control step. The test environment follows the same setup as policy testing (mean of all ensemble members). The resulting performance and computation times are shown in Table \ref{tab:mpc_rotation} and Table \ref{tab:mpc_temperature}.


\begin{table}[H]
\centering
\caption{Rotation tracking: control-cycle time and evaluation return for the learned policies and sampling-based MPC. $K$ is the number of candidate action sequences, $H$ is the planning horizon, and CEM$\times$2 denotes two
cross-entropy-method iterations. Higher return is better.}
\label{tab:mpc_rotation}
\small
\begin{tabular}{@{}lrr@{}}
\toprule
Controller & Time (ms/cycle) & Return \\
\midrule
\textbf{Latent-conditioned policy (ours)} & \textbf{0.29} & $\mathbf{-298.5}$ (17.6) \\
MPC ($K{=}32, H{=}5$, CEM$\times$2) & 42.0 & $-304.0$ (18.5) \\
MPC ($K{=}16, H{=}2$, CEM$\times$2) & 14.9 & $-358.4$ (37.1) \\
MPC ($K{=}64, H{=}3$, CEM$\times$2) & 28.3 & $-366.0$ (26.2) \\
MPC ($K{=}96, H{=}2$, CEM$\times$2) & 22.4 & $-366.5$ (32.9) \\
MPC ($K{=}32, H{=}2$, CEM$\times$2) & 16.5 & $-371.7$ (31.1) \\
Feedforward policy & 0.08 & $-383.2$ (14.5) \\
MPC ($K{=}64, H{=}2$, CEM$\times$2) & 20.0 & $-393.0$ (34.0) \\
\bottomrule
\end{tabular}
\end{table}


\begin{table}[H]
\centering
\caption{Electron-temperature tracking: control-cycle time and evaluation
return for the learned policies and sampling-based MPC. $K$ is the number of
candidate action sequences, $H$ is the planning horizon, and CEM$\times$2
denotes two cross-entropy-method iterations. Higher return is better.}
\label{tab:mpc_temperature}
\small
\begin{tabular}{@{}lrr@{}}
\toprule
Controller & Time (ms/cycle) & Return \\
\midrule
\textbf{Latent-conditioned policy (ours)} & \textbf{0.29} & $\mathbf{-148.6}$ (10.9) \\
MPC ($K{=}32, H{=}5$, CEM$\times$2) & 42.0 & $-159.7$ (12.5) \\
MPC ($K{=}64, H{=}3$, CEM$\times$2) & 28.3 & $-168.4$ (14.9) \\
Feedforward policy & 0.08 & $-183.8$ (13.7) \\
MPC ($K{=}32, H{=}2$, CEM$\times$2) & 16.5 & $-198.5$ (16.5) \\
MPC ($K{=}64, H{=}2$, CEM$\times$2) & 20.0 & $-198.6$ (17.2) \\
MPC ($K{=}16, H{=}2$, CEM$\times$2) & 14.9 & $-203.8$ (17.2) \\
MPC ($K{=}96, H{=}2$, CEM$\times$2) & 22.4 & $-203.9$ (17.4) \\
\bottomrule
\end{tabular}
\end{table}

These results show the expected trade-off. Increasing the planning horizon and number of sampled trajectories improves MPC performance, but also substantially increases online computation. The best-performing MPC (using a compute of over 42 ms) exceeds the feedforward policy but remains below the latent conditioned policy. 

Importantly, under our implementation using four CPU threads, the best-performing MPC requires approximately 42 ms per control cycle, whereas our deployed latent-conditioned policy requires only 0.29 ms (over two orders of magnitude faster). In the DIII-D Plasma Control System, the control interval is 20 ms, of which approximately 8–11 ms are already consumed by diagnostic acquisition and RTCAKENN reconstruction. Consequently, while MPC planning improves as computational effort increases, it does not satisfy the available real-time budget in our deployment setting, whereas the learned policy does.



\end{document}